\documentclass[conference]{IEEEtran}
\IEEEoverridecommandlockouts
\usepackage{cite}
\usepackage{amsmath,amssymb,amsfonts}
\usepackage{algorithmic}
\usepackage{graphicx}
\usepackage{textcomp}
\usepackage{xcolor}
\usepackage{placeins}
\usepackage{multirow}

\def\BibTeX{{\rm B\kern-.05em{\sc i\kern-.025em b}\kern-.08em
    T\kern-.1667em\lower.7ex\hbox{E}\kern-.125emX}}

\begin{document}

\title{Trajectory-Aware Adaptive Inference in Object Detection Models\\
}

\author{\IEEEauthorblockN{Grigorios Papanikolaou$^{1}$, Ioannis Kontopoulos$^{1}$, Giannis Spiliopoulos$^{2}$, Dimitris Zissis$^{2}$, Konstantinos Tserpes$^{1}$}
\IEEEauthorblockA{\textit{$^{1}$Department of Electrical and Computer Engineering, National Technical University of Athens, Greece}\\
\textit{$^{2}$Department of Product and Systems Design Engineering, University of the Aegean, Syros, Greece
}}
}

\maketitle

\begin{abstract}
The increasing integration of sensors in autonomous maritime navigation has led to large-scale multimodal datasets, raising challenges in achieving efficient real-time perception. In such systems, object detection and trajectory perception of nearby vessels are tightly coupled, particularly in dynamic environments such as maritime navigation. However, the efficiency of object detection models during inference remains an often-overlooked aspect. To this end, we build upon an existing object detection framework by incorporating GPS trajectory data into the inference process to enable input-adaptive computation. Specifically, we introduce an early-exit mechanism in a YOLOv8-based detector that incorporates motion cues - such as inter-vessel distances. Frames of vessels that are separated by short distances, converging with high speed, are processed using the full model, while only a subset of the network's architecture is activated otherwise. The difficulty degree (or scene complexity) of a frame or set of frames per second is evaluated by leveraging inter-object distance and the rate at which the distance between them decreases. Experimental results demonstrate that this strategy maintains satisfactory detection performance while significantly reducing inference time and computational cost, thus enabling a flexible trade-off between accuracy and efficiency compared to full-model inference.

\end{abstract}


\section{Introduction}


The multitude of mobility data from possibly heterogeneous sensors is not effectively leveraged due to its fragmented and usually incomplete nature \cite{nuscenes}\cite{qian_review_2025}. Trajectory data collected from various sensors are difficult to handle because of the complexity in the fusion and pre-processing stages, limiting their value towards downstream analysis \cite{liu_real_2023}. In addition, another problem pertains to the fundamental role of robust perception in autonomous systems because there is a need for safe and reliable operation in real-world conditions where minor errors can lead to undesirable and severe consequences \cite{feng_deep}. Therefore, effectively handling and incorporating trajectory data from multiple sensors can probably aid towards achieving critical goals such as precise and real-time perception \cite{feng_deep}. Furthermore, real-time scenarios impose constraints in latency, computational efficiency, and the possible trade-off between accuracy and limited resource management \cite{redmon_you} \cite{wang_survey_2025} further complicating the handling of trajectories. Within this context, the role of object detection is important as it allows for accurate and real-time identification and localization, helping in domains such as trajectory interpretation and modeling of interactions \cite{wang_survey_2025} \cite{yaseen_what_2024}.


In use cases revolving around Autonomous Sea Vehicles (ASVs) under low-latency and energy-efficient computation constraints, real-time perception is required. In this context, object detection is important for tracking nearby vessels, avoiding collision, and situational awareness in general. However, popular detectors such as YOLOv8 demand the same computational effort on all inputs, invariant to their objective difficulty \cite{yaseen_what_2024}. The models of the YOLO family are non-adaptive during inference, utilizing their full computational graph, and this is not optimal for resource-constrained environments. There are various approaches to prevent this phenomenon from occurring, such as distillation \cite{hinton2015distilling}, quantization \cite{jacob2018quantization}, and pruning \cite{han2015deep}, as well as multi-exit neural networks \cite{teerapittayanon2016branchynet}. To be more precise, multi-exit neural networks are able to selectively activate sections of their computational graph based on internal signals quantifying the models certainty. The internal signals mentioned are based on entropy or max probability and they are not explicitly based on external context, such as trajectory data and motion \cite{teerapittayanon2016branchynet}. The exits of multi-exit neural networks, which denote the points in the neural network's architecture where the inference will stop, are often optimized jointly, and usually each one of them has unique strengths and weaknesses \cite{10.1145/3698767}. For example, in Convolutional Neural Networks (CNNs) shallow-depth exits are able to make predictions based on low-level features, whereas deeper exits are basing their predictions on high-level features \cite{10.1145/3698767}. 


In the context of ASVs, there are multi-modal data provided through cameras and GPS trajectories. The hypothesis of this work is based on the assumption that scene difficulty is correlated with object proximity and relative motion, in addition to vision not being sufficient for quantifying these metrics fully. For example, one important mobility cue to mention is closure rate, defined as the rate at which the in-between distance of moving objects changes over time within a scene. Therefore, incorporation of trajectory-based signals into the inference procedure would enable input-adaptive computation, resulting in reduced latency and energy consumption whilst preserving precise predictive performance. In this work, instead of relying on confidence-based early exiting, we leverage external trajectory-derived cues to determine which part of the computational graph should be activated. This enables a more informed and context-aware inference strategy, particularly suited to dynamic environments such as maritime scenes.


Specifically, a YOLOv8 detector is leveraged, and its three detection heads used across different scales are treated as implicit exits. Initially, there is a scale-aware fine-tuning procedure by assigning a different learning rate to each YOLOv8 detection head. Subsequently, the model utilizes and quantifies the trajectory-based difficulty using signals such as the Haversine distance or closure rate between two moving objects being tracked within the scene. The final policy being proposed operates by leveraging only a subset of the heads for object detection, whilst activating all the detection heads in difficult frames. The contributions of this study can be summarized as follows:

\begin{itemize} 
    \item The combination of mobility data and vision in a multi-modal adaptive inference framework.
    
    \item The scale-aware fine-tuning methodology focusing in the learning rate modulation of individual heads.
    
    \item The empirical evaluation of the proposed YOLOv8 model provides insights on the per-head projected latency-accuracy trade-offs and computational cost savings. In addition, the effectiveness of the proposed adaptive inference criterion in terms of latency is empirically validated. 
    
\end{itemize}

\section{Related Work}

Adaptive inference is a field that has been studied extensively aiming to reduce the computational cost of inference by dynamically adjusting the utilization of a model's computational graph based on input difficulty. Multi-exit neural networks leverage intermediate classifiers along the main classifier and paradigms such as BranchyNet \cite{teerapittayanon2016branchynet} and Shallow-Deep Networks \cite{kaya2019shallow} allow predictions to be made prematurely. There are various measures of quantifying the confidence of a model as a signal of processing termination and the most popular are entropy and maximum softmax probability \cite{10.1145/3698767}. These methods are effective but there is an inherent limitation based on the observation that they rely on model confidence not external contextual information.

Efficient object detection has been a central topic of interest in recent research. One of the most established family of architectures is YOLO \cite{redmon_you} \cite{yaseen_what_2024} and is based on multi-scale detection and optimized network design \cite{redmon_you}. Over the years many variations of this architecture have been proposed aiming to improve the predictive performance and efficient inference trade-off. This fact along with complementary techniques such as compression and knowledge distillation have led to significant progress towards efficient inference in this domain \cite{hinton2015distilling} \cite{jacob2018quantization} \cite{han2015deep}. The efficiency gains may be substantial but they refer to static reductions in computation and latency regardless of input difficulty.

Perception methods based on multiple modalities \cite{qian_review_2025} \cite{feng_deep} combine data from sensors such as cameras and GPS to improve detection accuracy and robustness for systems deployed in complex environments \cite{qian_review_2025}. Research focusing in architectures leveraging fusion in combination with large-scale datasets \cite{nuscenes} indicates that integration of heterogeneous inputs enhances scene understanding and performance in object detection. Existing approaches are focused in improving predictive performance through optimization in feature-level or decision-level fusion \cite{qian_review_2025} \cite{feng_deep}. Leveraging auxiliary modalities towards guiding computational resource allocation during inference is not as explored in comparison.

Representations based on trajectories or mobility data have been explored in the context of tracking and modeling interactions \cite{wang_survey_2025}. This can be leveraged in capturing dynamic relationships or predicting future states \cite{wang_survey_2025}. The aforementioned practices exploit the use of temporal and spatial dependencies to improve forecasting accuracy as well as behavioral understanding \cite{wang_survey_2025}. This study utilizes trajectory-based motion signals, such as inter-object distance and closure rate towards adaptive inference. The proposed object detection framework enables input-dependent computation by balancing latency whilst preserving an acceptable degradation of predictive performance.

\section{Methodology}

We propose a trajectory-aware adaptive inference framework that integrates motion information into an object detector's inference process. The pipeline consists of three main stages: detector adaptation, head-wise analysis, and trajectory-driven inference control. First, the base detector is fine-tuned on the target domain and its corresponding classes, taking into account the distribution of bounding box sizes in the training data. This step ensures that each detection head is appropriately calibrated for the downstream task. Following fine-tuning, we evaluate the performance, latency, and computational cost associated with each detection head independently, enabling a clear understanding of their individual contributions in the inference process. Second, vessel trajectory data is analyzed to extract motion cues, such as inter-vessel distance and the rate at which the distance between the objects within the scene changes (closure rate). These signals are used to estimate the difficulty of each frame. Frames involving closely spaced vessels or high-speed convergence are considered more challenging, while simpler scenarios correspond to well-separated or slowly moving objects. Finally, this difficulty estimate is used to guide inference, i.e., only a subset of detection heads is activated for low-complexity frames, whereas the full model is employed when higher precision is required. In this way, the framework achieves a dynamic trade-off between computational efficiency and detection performance. The first and second stages of the proposed framework towards trajectory-aware adaptive inference are theoretically grounded and justified by the insights offered in Section~\ref{par:layer_wise_training}. The third stage of the framework is further discussed in Section~\ref{par:criterion}.

\subsection{Multi-Exit Neural Networks}

Multi-exit neural networks have been widely explored as a means of enabling adaptive inference, particularly in computer vision and natural language processing. These architectures introduce intermediate prediction branches that allow early termination of inference, thereby reducing latency and computational cost while maintaining satisfactory accuracy. In the context of this work, we observe that YOLOv8 inherently exhibits a multi-exit structure. Specifically, it consists of three distinct components. The first component, being a Convolutional Neural Network, namely the \textbf{Backbone}, is responsible for extracting hierarchical feature maps, and can capture both low-level texture features as well as high-level semantic information. The second component is the \textbf{Neck} and is leveraged towards the refinement and fusion of the hierarchical features extracted from the backbone. The \textbf{Heads} (P3, P4, P5) are the last category of component, and they are responsible for generating the final predictions. Each head operates at a different spatial resolution and specializes in detecting objects of different sizes. The P3 head targets small objects using high-resolution feature maps, while P5 focuses on larger objects using lower-resolution representations. This architectural property allows us to reinterpret YOLOv8 as an implicit multi-exit network without introducing additional branches.



\subsection{Layer-wise Learning Rates Tuning}\label{par:layer_wise_training}

The pretrained YOLOv8 model is originally optimized on the COCO dataset, whose distribution of the objects' bounding box sizes may differ significantly from that of the target maritime domain. Since each detection head (P3, P4, P5) is responsible for a specific range of object sizes, such a distribution shift directly impacts their relative importance during fine-tuning. To account for this, we first analyze the distribution of bounding boxes sizes in the target dataset. Each object instance is categorized as small, medium, or large based on the geometric mean of the bounding box:
\begin{equation}
    m = \sqrt{width*height}
\end{equation}

Then, we assign different learning rates to each detection head. Heads corresponding to over-represented object sizes in the target dataset are assigned higher learning rates, allowing them to adapt more aggressively to the new domain. Conversely, heads associated with under-represented sizes receive lower learning rates to prevent overfitting. This strategy is particularly important for the P3 head, which operates on high-resolution feature maps and is more sensitive to variations in small-object density. By aligning the optimization dynamics with the dataset’s size characteristics, we ensure that each head is appropriately specialized, thereby improving both standalone performance and the effectiveness of the proposed early-exit mechanism.



\subsection{Trajectory-Aware Exiting Criterion}\label{par:criterion}

A central component of the proposed methodology is the use of trajectory data to guide adaptive inference. In this work, the decision criterion is derived exclusively from the motion of detected vessels, without explicitly modeling the movement or positioning of the observer. Although this prevents a direct spatial mapping between GPS and image coordinates, trajectory data -- when synchronized with video frames -- provides valuable contextual information about scene dynamics. Given a trajectory sampling of 1 data point per second, it is possible to estimate both the pairwise distances between vessels within a frame and their relative motion directly from latitude and longitude measurements taken from their GPS sensors. Pairwise distances are computed using the Haversine formula, while motion dynamics are captured through the closure rate. These two quantities jointly serve as indicators of frame difficulty. Scenarios where vessels are both spatially close and rapidly converging are considered more challenging, as they often involve increased interaction, potential occlusions in the frame, or critical navigation events. In contrast, frames with well-separated vessels and low relative motion are treated as simpler cases.

To exploit this distinction, we define a trajectory-aware exiting criterion that dynamically controls which detection heads of the model are activated during inference. The selection of candidate heads is informed both by motion cues and by prior analysis of the dataset’s distribution of bounding box sizes, which identifies the most frequently utilized heads under typical conditions. The head selection policy is defined as:

\begin{equation}
\mathcal{H}_t =
\begin{cases}
\{P3\}, & \text{if } d_t > \tau_1 \;\land\; v_t < \tau_2 \\[4pt]
\{P3, P4, P5\}, & \text{otherwise}
\end{cases}
\label{eq:head_policy}
\end{equation}

where $d_t$ denotes the pairwise distance of objects within the scene or frame between vessels at time $t$, $v_t$ represents the corresponding closure rate at time $t$, and $\tau_1$ and $\tau_2$ denote the respective thresholds. When vessels are sufficiently distant and exhibit low convergence, only the high-resolution detection head (P3) is activated, enabling reduced computational cost and latency. Otherwise, the full computational graph of the detector is utilized, which contains all of the heads, thus preserving detection performance in more complex scenarios. In practice, frame difficulty is evaluated at a per-second granularity. Every second, the $d_t$ and $v_t$ values are calculated, taking into account the frames and trajectory points within that one-second window.

\section{Experiments}

\subsection{Dataset}
The dataset used in this work comprises two synchronized modalities: (i) video recordings captured by onboard cameras mounted on custom Autonomous Surface Vehicles (ASVs), and (ii) trajectory data obtained from the onboard GPS sensors of the same platforms. The videos depict maritime scenes at sea level, including the motion of two ASVs as well as additional vessels present in the background. Notably, the camera itself is in motion, as it is mounted on an ASV, introducing additional variability in the visual data. The accompanying trajectory data is temporally aligned with the video streams and provides precise information about ASV motion in real-world geographic coordinates. This information is leveraged to extract motion cues, such as inter-vessel distance and relative velocity (closure rate), which are subsequently used to assess the difficulty of each frame. As such, the dataset enables the joint exploitation of visual and trajectory information, forming the basis for the proposed trajectory-aware adaptive inference framework. Finally, for training and fine-tuning the object detection models, a separate annotated dataset is utilized, consisting of frames containing ASVs and boats with bounding box annotations for two classes: ASV and Boat. The Boat class refers to vessels that remain static in the background, while the ASV class captures the actively moving platforms. In total, the dataset consists of 250 records of GPS data (125 per ASV) and 3686 video frames of a 125-second video. The ASV platforms were developed by the SmartMove\footnote{https://smartmove.aegean.gr/} lab, which is part of the Department of Product and Systems Design Engineering of the University of the Aegean. The recordings of both the videos and the GPS data took place in a dedicated maritime space of the SmartMove lab.

\subsection{Object Detection models} 

YOLOv8 was chosen instead of other popular options, such as YOLOv11, due to its wide adoption in industrial use cases and optimized AI accelerators. Furthermore, other YOLO versions, such as v11, employ attention mechanisms that require bigger and more diverse datasets.

\begin{itemize} 
    \item \textbf{YOLOv8 Nano :} This is the most lightweight model of the YOLOv8 family (3.16 million parameters) and is ideal for use in environments with limited resources. For example, it is ideal for use cases such as edge deployment in real-time detection of objects regarding Autonomous Surface Vehicles (ASVs).
    
    \item \textbf{YOLOv8 Small :} This model is ideal for inference both in CPUs and GPUs balancing speed and accuracy (11.17 million parameters). The utilization of an improved Path Aggregation Network (PAN) along with enhanced spatial pyramid pooling leads to improved feature fusion and accuracy.
    
    \item \textbf{YOLOv8 Medium :} This architecture represents the mid-tier of YOLOv8 family  (25.9 million parameters). In comparison to the two aforementioned architectures it features a deeper backbone and neck component and is therefore performing better across various datasets and tasks.

\end{itemize}

\subsection{Training Regime \& Learning Rates}

\begin{table*}[!t]
\centering
\caption{Scale composition of the training set and resulting per-component learning rates.
Normalized ratios $r_k$ are computed relative to the dominant category (medium).
Base learning rate $\eta_0 = 10^{-3}$.}
\label{tab:lr_schedule}
\begin{tabular}{|l|c|c|c|c|c|c|}
\hline
\textbf{Component} & \textbf{Size range} & \textbf{Instances} & $f_k$ (\%) & $\omega_k$ & $r_k$ & $\eta_{P_k}$ \\
\hline
Small (P3)  & $s < 32$px          & 316 & 20.93 & 1.5 & 0.618 & $6.18 \times 10^{-4}$ \\
\hline
Medium (P4) & $32 \leq s < 96$px  & 767 & 50.79 & 1.0 & 1.000 & $1.00 \times 10^{-3}$ \\
\hline
Large (P5)  & $s \geq 96$px       & 427 & 28.28 & 0.7 & 0.390 & $3.90 \times 10^{-4}$ \\
\hline
Neck        & ---                 & --- & ---   & --- & 0.800 (fixed) & $8.00 \times 10^{-4}$ \\
\hline
\end{tabular}
\end{table*}

During fine-tuning, we adopt a scale-aware learning rate assignment strategy to account for the object size distribution in the training split. Specifically, each annotated bounding box is categorized into a scale group (small, medium, large) based on the geometric mean of its pixel dimensions $s$, using predefined thresholds as shown in Table~\ref{tab:lr_schedule}. To reflect the varying difficulty associated with detecting objects of different sizes, a weight $\omega_{k}$ is assigned to each scale category, with higher values corresponding to more challenging cases (e.g., small objects). For each category, a weighted fraction score is computed as $\omega_{k}f_{k}$, where $f_k$ denotes the relative frequency of instances in that category. These scores are then normalized with respect to the maximum across all categories:


\begin{equation}
r_{k} = \frac{\omega_{k}f_{k}}{\max_{j}(\omega_{j}f_{j})}
\label{eq:rate}
\end{equation}

The resulting normalized factors $r_k$ are used to modulate the learning rates of the corresponding detection heads. The head associated with the highest score $r_k = 1$ is assigned the base learning rate $\eta_0$, while the remaining heads receive proportionally scaled learning rates. This ensures that heads responsible for more prominent or challenging object scales adapt more aggressively during fine-tuning. In contrast, the neck component is trained using a fixed learning rate of $0.8\,\eta_0$, independent of the scale distribution. The final scale composition and corresponding learning rates are summarized in Table~\ref{tab:lr_schedule}. In our setting, the P4 head receives the highest learning rate ($1.00 \times 10^{-3}$), followed by P3 ($6.18 \times 10^{-4}$) and P5 ($3.90 \times 10^{-4}$), reflecting the predominance of medium-scale objects in the dataset.


\subsection{Experimental Evaluation}

This section evaluates the feasibility of adaptive inference through selective activation of detection heads. The goal is to determine whether individual heads can provide sufficient detection performance while yielding measurable reductions in computational cost and inference latency. To this end, we evaluate each detection head (P3, P4, P5) independently across multiple YOLOv8 variants on the validation split used during fine-tuning. Detections are attributed to heads based on their originating feature maps, while computational savings and inference speedups in comparison to leveraging the full computational graph are estimated by profiling the corresponding subgraphs required for isolated head execution. Table~\ref{tab:head_results} indicates the distribution of detections across heads, along with the corresponding inference speedups and computational savings achieved when each head is used independently, while Table~\ref{tab:head_performance} reports the standalone detection performance of each head in terms of mAP@50, precision, and recall.

\begin{table*}[!t]
\centering
\caption{Model efficiency comparison against full path across detection scales (P3, P4, P5)}
\begin{tabular}{|l|c|c|c|c|c|c|c|c|c|c|}
\hline
\multirow{2}{*}{\textbf{Model}} & \multirow{2}{*}{\textbf{Total}} & \multicolumn{3}{c|}{\textbf{Detections}} & \multicolumn{3}{c|}{\textbf{Speedup}} & \multicolumn{3}{c|}{\textbf{FLOPs Savings (\%)}} \\
\cline{3-11}
 &  & P3 & P4 & P5 & P3 & P4 & P5 & P3 & P4 & P5 \\
\hline
YOLOv8 Nano   & 590 & 168 & 277 & 145 & $\times$1.61 & $\times$1.45 & $\times$1.34 & 25.08 & 33.79 & 32.71 \\
\hline
YOLOv8 Small  & 602 & 212 & 286 & 104 & $\times$1.56 & $\times$1.53 & $\times$1.27 & 24.41 & 28.54 & 25.45 \\
\hline
YOLOv8 Medium & 556 & 228 & 309 & 19  & $\times$1.31 & $\times$1.33 & $\times$1.17 & 21.26 & 20.56 & 18.33 \\
\hline
\end{tabular}
\label{tab:head_results}
\end{table*}

The results, as demonstrated in  Tables~\ref{tab:head_results} and \ref{tab:head_performance}, reveal a consistent pattern across all model variants. The P4 head dominates in terms of the number of detections and achieves the highest standalone performance, indicating that the dataset is primarily composed of medium-scale objects. In contrast, the contribution of the P5 head decreases significantly as model capacity increases, suggesting redundancy in large-scale feature representations. At the same time, detections from the P3 head increase with model scaling, reflecting improved sensitivity to smaller objects in larger models.

From a computational perspective, the FLOPs savings reported in Table~\ref{tab:head_results} highlight that activating individual heads leads to non-uniform efficiency gains. These differences arise from the architecture of YOLOv8, where model scaling increases both depth and width, causing the backbone to dominate the overall computational cost in larger variants. As a result, the relative contribution of the detection heads becomes less significant, leading to a more compressed distribution of computational savings—particularly in the YOLOv8 Medium model. Furthermore, the bidirectional structure of the neck introduces asymmetries in the computational graph, explaining why FLOPs reductions do not translate linearly into latency improvements.

\begin{table*}[!t]
\centering
\caption{Model performance comparison across detection scales (P3, P4, P5)}
\begin{tabular}{|l|c|c|c|c|c|c|c|c|c|c|}
\hline
\multirow{2}{*}{\textbf{Model}} 
& \multicolumn{3}{c|}{\textbf{mAP@50}} 
& \multicolumn{3}{c|}{\textbf{Precision}} 
& \multicolumn{3}{c|}{\textbf{Recall}} \\
\cline{2-10}
& P3 & P4 & P5 
& P3 & P4 & P5 
& P3 & P4 & P5 \\
\hline

YOLOv8 Nano   
& 0.6179 & 0.7959 & 0.6709 
& 0.8121 & 0.8960 & 0.9786 
& 0.3818 & 0.6461 & 0.3540 \\
\hline

YOLOv8 Small  
& 0.6736 & 0.8084 & 0.5903 
& 0.8215 & 0.8980 & 0.9167 
& 0.4805 & 0.6645 & 0.2458 \\
\hline

YOLOv8 Medium 
& 0.7235 & 0.8713 & 0.5112 
& 0.8506 & 0.9346 & 0.9643 
& 0.5411 & 0.7632 & 0.0560 \\
\hline

\end{tabular}
\label{tab:head_performance}
\end{table*}

The inference time results further support these observations. The P3 head consistently provides the highest speedup, making it well-suited for low-complexity scenarios. The P4 head remains competitive, particularly in smaller model variants, while the P5 head offers limited latency benefits. As model size increases, the dominance of the backbone reduces the relative gains achieved by selective head activation, mirroring the trends observed in FLOPs savings.

Finally, the standalone performance metrics in Table~\ref{tab:head_performance} demonstrate that individual heads can operate as reliable predictors under appropriate conditions. The P4 head consistently achieves the highest mAP@50 and precision across all variants, reinforcing its central role in the detection pipeline. Across all heads, a consistent gap between precision and recall is observed, with significantly higher precision values. This indicates that individual heads tend to produce conservative predictions, favoring high-confidence detections at the expense of recall.

\begin{figure*}[!t]
    \centering
    \includegraphics[width=\columnwidth]{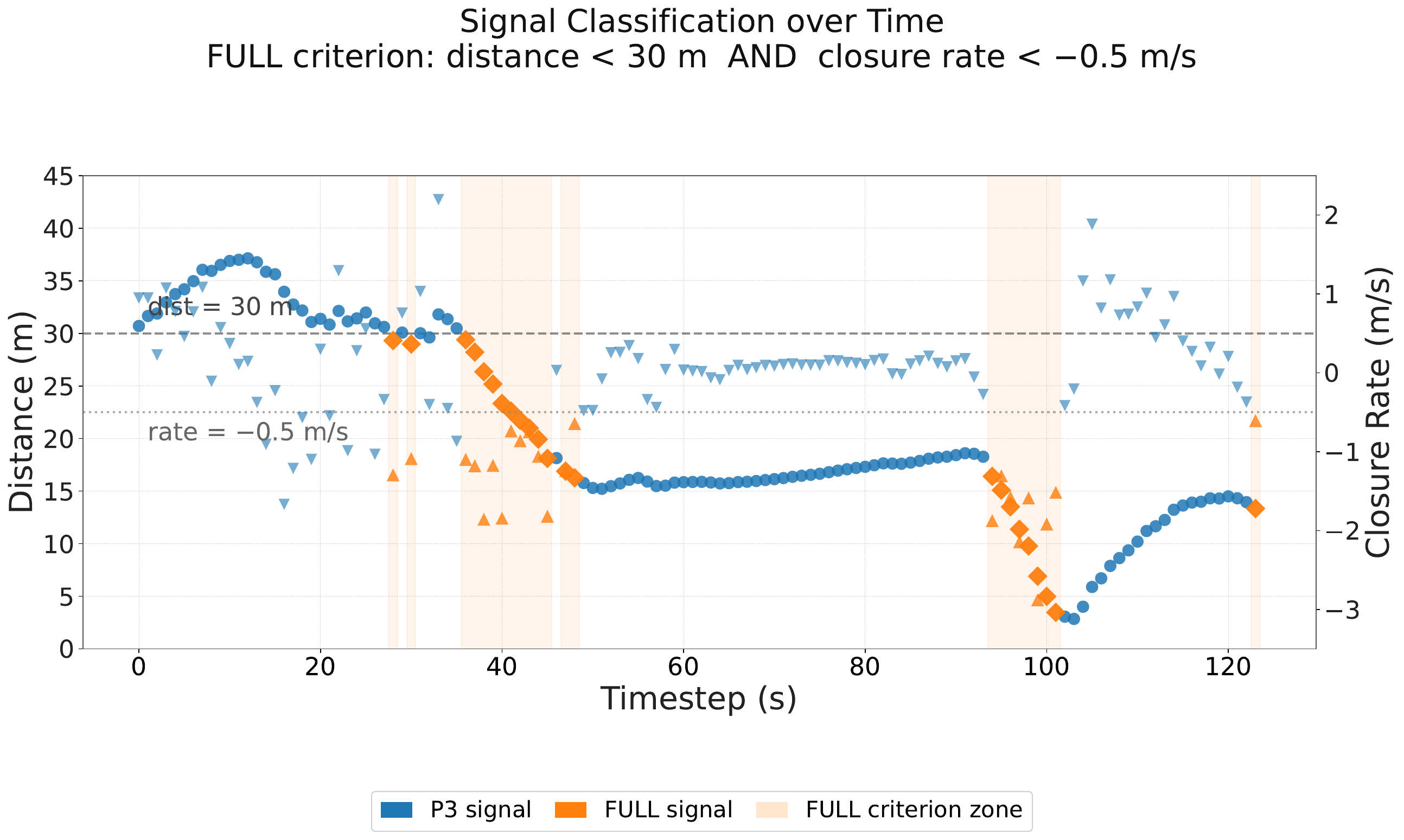}
    \caption{This figure presents the changes between the usage of heads based on the signal derived from trajectories.}
    \label{fig:mreg}
\end{figure*}

To further validate the proposed adaptive inference strategy, we conduct additional experiments in a realistic object detection scenario. Building upon the insights of the previous evaluation, this experiment investigates the effectiveness of trajectory-aware head selection in a dynamic maritime setting. Specifically, a YOLOv8 model is deployed to detect two autonomous surface vehicles in a short, sea-level video captured from a moving ASV. The model is trained to detect both ASVs and boats, enabling real-time tracking under resource-constrained conditions. The objective is to assess whether trajectory-aware indicators -- namely, inter-vessel distance and closure rate -- can effectively guide the selection between partial and full model inference. The exiting criterion is applied using empirically selected thresholds, designed to balance computational savings with sufficient utilization of multiple heads. In particular, the distance threshold is set to 30 meters ($\tau_1$), while the closure rate threshold is set to $0.5~m/s$ ($\tau_2$). Frames satisfying the low-complexity condition are processed using only the P3 head, whereas more challenging frames trigger the use of the full model. Figure~\ref{fig:mreg} illustrates, for each trajectory timestamp, the corresponding closure rate and distance between the two moving ASVs, along with the selected heads.The results, demonstrate that the trajectory-aware signal effectively modulates head usage over time. Out of a total of $3,686$ frames, $3,027$ are processed using only the P3 head, while $659$ frames are evaluated using the full model. This distribution confirms that the majority of frames correspond to low-complexity scenarios, where reduced computation is sufficient. In terms of efficiency, the median inference latency is reduced from $10.097~ms/frame$ when using the full model to $6.686~ms/frame$ when using only the P3 head. Consequently, the total processing time decreases from a projected $37.22$ seconds (full model) to $26.89$ seconds with the proposed adaptive framework. These results demonstrate that trajectory-aware head selection can significantly reduce inference time while maintaining effective tracking performance.

Based on the experimental results, there are a few crucial conclusions to be drawn. The first pertains to the fact that the proposed framework can significantly lower the processing overhead. In addition, the per-head evaluation experiments, based on profiling through intermediate hooks and tracing of the computational graph, provided insights into each head's individual performance and projected savings, in comparison to leveraging the full model. The proposed fine-tuning strategy focusing on bounding box size distribution has yielded tangible results, as demonstrated by the evaluation of the P4 head.

\section{Conclusions and Future Work}

In this work, we presented a trajectory-aware early-exiting mechanism that dynamically controls detection head activation in a YOLOv8-based architecture using inter-object distance and closure rate. Experimental results show that individual heads can serve as reliable predictors, enabling significant reductions in computational cost and inference latency while maintaining comparable detection performance. This highlights the suitability of the approach for real-time, resource-constrained applications such as autonomous maritime navigation.

Future work will explore the generalization of this method to alternative architectures, such as DETR, as well as its robustness under challenging conditions, including adverse weather, sensor noise, and low visibility. Additionally, incorporating explicit mappings between GPS and image coordinates is expected to further enhance the integration of trajectory and visual information in real-world deployments.


\section{Acknowledgments}
This work was supported by the MUSIT Project through the European Union’s Horizon Europe Framework Programme (HORIZON), under Marie Sklodowska-Curie grant agreement no. 101182585. The work only reflects the authors’ views; the EU Agency is not responsible for any use of the information it contains. 

\bibliographystyle{IEEEtran}
\bibliography{references}
\end{document}